\documentclass{article}

\usepackage{microtype}
\usepackage{graphicx}
\usepackage{subcaption}
\usepackage{booktabs}
\usepackage{hyperref}

\usepackage[accepted]{icml2026}

\usepackage{amsmath}
\usepackage{amssymb}
\usepackage{mathtools}
\usepackage{amsthm}
\usepackage{xcolor}

\usepackage[capitalize,noabbrev]{cleveref}

\theoremstyle{plain}

\icmltitlerunning{Library Drift: A Failure Mode in Self-Evolving Agents}

\begin{document}

\twocolumn[
\icmltitle{Library Drift: Diagnosing and Fixing a Silent Failure Mode\\in Self-Evolving LLM Skill Libraries}

\icmlsetsymbol{equal}{*}

\begin{icmlauthorlist}
\icmlauthor{Xing Zhang}{aws}
\icmlauthor{Yanwei Cui}{aws}
\icmlauthor{Guanghui Wang}{aws}
\icmlauthor{Ziyuan Li}{hsbc}
\icmlauthor{Wei Qiu}{hsbc}
\icmlauthor{Bing Zhu}{hsbc}
\icmlauthor{Peiyang He}{aws}
\end{icmlauthorlist}

\icmlaffiliation{aws}{AWS Generative AI Innovation Center}
\icmlaffiliation{hsbc}{HSBC Holdings Plc., HSBC Technology Center, China}

\icmlcorrespondingauthor{Peiyang He}{peiyan@amazon.com}

\icmlkeywords{LLM Agents, Self-Evolving Skills, Failure Modes, Library Drift, Lifecycle Management}

\vskip 0.3in
]

\printAffiliationsAndNotice{}

\begin{abstract}
Self-evolving skill libraries face a silent failure mode we term
\emph{library drift}: unbounded skill accumulation without
outcome-driven lifecycle management causes retrieval degradation,
false-positive injections, and performance stagnation. Recent
evaluation confirms the symptom (LLM-authored skills deliver
+0.0pp gain while human-curated ones deliver
+16.2pp; SkillsBench~\citep{li2026skillsbench}), yet the underlying
mechanism has not been isolated.
We provide (1)~a \textbf{reproducible trigger}: ablations that
isolate drift: one disables skill injection (flat floor,
+0.002), one imposes premature retirement (active harm,
$-$0.019); (2)~\textbf{trace-level
diagnostics}: an append-only evidence log with per-skill
contribution scores, attribution verdicts, and router engagement
metrics that make the failure visible before it reaches end-task
scores; and (3)~a \textbf{verified fix}: a minimal governance
recipe (outcome-driven retirement + bounded active-cap + meta-skill
authoring prior) that lifts held-out pass@1 from a 0.258 baseline
to a late-window mean of 0.584 (rolling gain $+$0.328) on MBPP+
hard-100 over 100 rounds. Eight ablations
decompose which governance mechanisms are load-bearing and which
are subsumed, providing a concrete playbook for diagnosing library
drift in any self-evolving agent.
\end{abstract}

\section{Introduction}
\label{sec:intro}

Self-evolving skill libraries, pioneered by
Voyager~\citep{wang2023voyager}, let frozen LLM agents accumulate
reusable procedural knowledge without weight updates. The promise is
compounding: each solved task deposits a skill that accelerates
future tasks. Yet SkillsBench~\citep{li2026skillsbench} reveals a
striking gap: LLM-authored skills deliver +0.0pp over no-skill
baselines while human-curated ones deliver +16.2pp. A recent
survey~\citep{zhang2026ecs} of 20+ such systems reports that
lifecycle management (versioning, conflict detection,
deprecation) is ``largely neglected.'' The libraries grow, but
the agents do not improve.

We identify a specific, diagnosable failure mode behind this gap:
\textbf{library drift}. As skills accumulate without quality gates,
the library degrades retrieval precision, injects stale or harmful
guidance, and the agent's effective performance stagnates or drops
below its no-skill baseline. The failure is \emph{silent}: end-task
metrics decline gradually with no explicit error signal, making
trace-level diagnostics essential. While we demonstrate drift on
single-call code tasks, the mechanism applies more acutely to
multi-step agents, where a single misleading injection compounds
through downstream tool calls and planning decisions.

This paper makes three contributions toward understanding and
fixing failure modes in agentic systems:

\begin{enumerate}
  \item \textbf{Failure definition and reproducible trigger}
  (Sec.~\ref{sec:failure}): We define library drift operationally
  and show two ablations that bracket it: one establishes the
  no-skill floor (A1, +0.002), one triggers active harm via
  premature retirement (A4, $-$0.019), providing minimal
  reproductions for any system.

  \item \textbf{Trace-level diagnostics beyond final scores}
  (Sec.~\ref{sec:diagnostics}): An append-only evidence log with
  per-skill contribution scores, attribution verdicts, and router
  engagement metrics exposes drift at per-skill granularity, before
  it reaches aggregate pass@1.

  \item \textbf{Verified fix with documented trade-offs}
  (Sec.~\ref{sec:fix}): A minimal governance recipe
  (Ratchet\footnote[2]{Implementation, diagnostics, and ablation harness:
  \url{https://github.com/amazon-science/Self-Evolving-Agents-Ratchet}}):
  outcome-driven retirement, bounded active-cap,
  meta-skill authoring prior, lifts held-out pass@1 by +0.328.
  Eight ablations decompose which mechanisms are load-bearing and
  which are safely removed.
\end{enumerate}

\begin{figure*}[t]
  \centering
  \includegraphics[width=\textwidth]{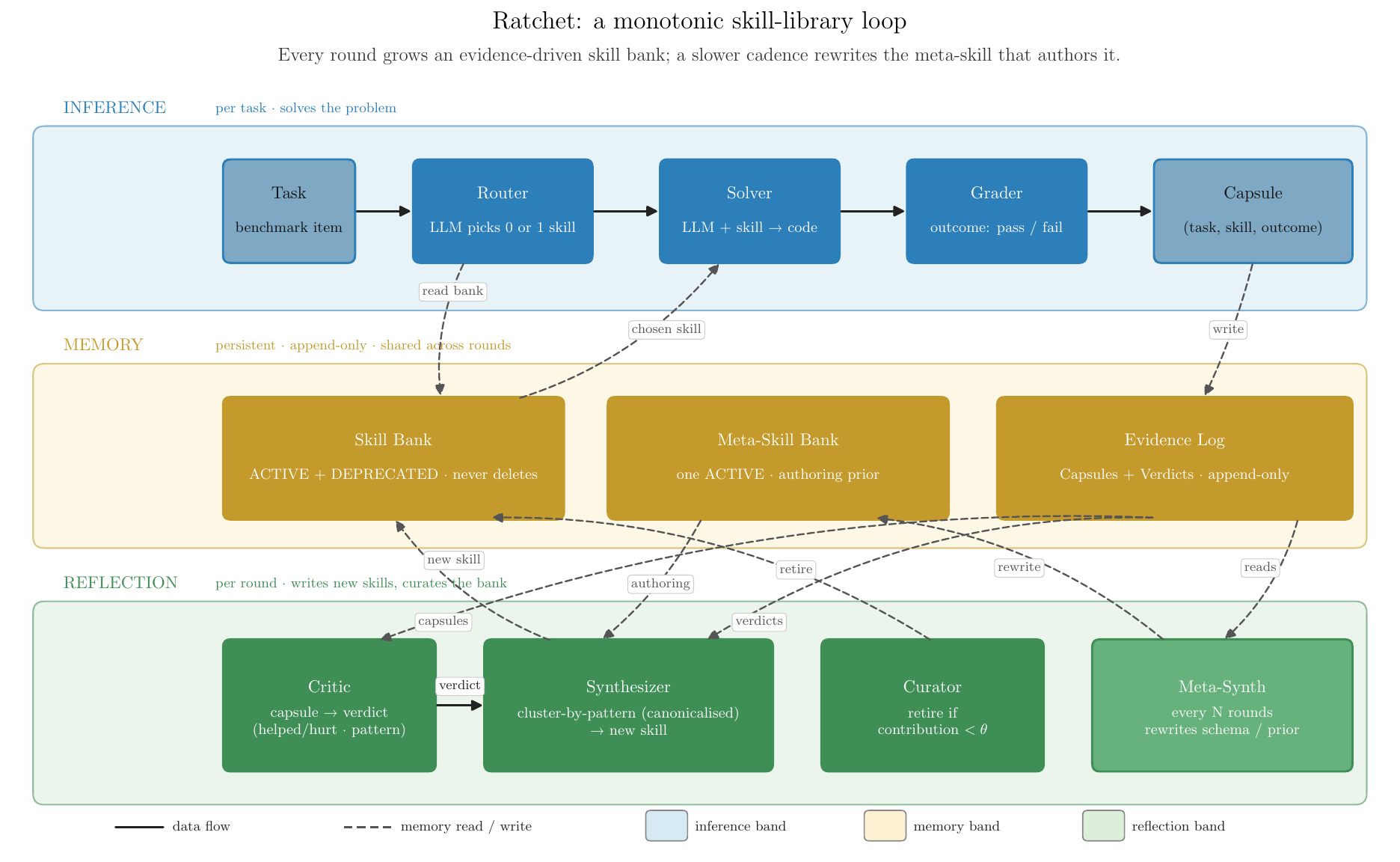}
  \caption{The Ratchet loop and where library drift is diagnosed and
  fixed. \textbf{Inference} (top): each task flows through
  Router\,$\to$\,Solver\,$\to$\,Grader\,$\to$\,Capsule.
  \textbf{Memory} (middle): Skill Bank, Meta-Skill, and Evidence Log.
  \textbf{Reflection} (bottom): the Critic produces attribution
  verdicts (the diagnostic signal); the Curator retires
  under-performers and enforces the bounded cap (the fix). Without
  outcome-driven retirement and the bounded cap, the Skill Bank
  accumulates unchecked and library drift emerges.}
  \label{fig:hero}
\end{figure*}

\section{Background and Related Work}
\label{sec:related}

\paragraph{Self-evolving skill libraries.}
Voyager~\citep{wang2023voyager} pioneered the ever-growing
executable skill library in Minecraft;
ExpeL~\citep{zhao2024expel} extracts textual insights from
trajectories but lacks outcome-driven retirement;
AutoManual~\citep{chen2024automanual} compiles instruction manuals;
CASCADE~\citep{huang2025cascade} pairs meta-skills with cumulative
creation; AutoSkill~\citep{yang2026autoskill} adds versioning but
does not retire on measured per-task contribution.
Concurrent work advances orthogonal axes:
EvolveR~\citep{wu2025evolver} distills trajectories into strategic
principles; Trace2Skill~\citep{alibaba2026trace2skill} induces
skills from trace pools; Strategy~Genes~\citep{wang2026strategygenes}
attaches failure history as metadata.
A recent survey~\citep{zhang2026ecs} of 20+ such systems reports
that lifecycle management is ``largely neglected''; none pairs
outcome-driven retirement with a bounded active-cap, the two
mechanisms our ablation analysis identifies as load-bearing for
preventing drift.

\paragraph{Agent failure modes.}
Prior work catalogs failure modes in tool
use~\citep{schick2023toolformer}, planning~\citep{yao2023react},
and self-correction~\citep{shinn2023reflexion,
madaan2023selfrefine}. Library drift is a distinct failure mode
specific to \emph{persistent cross-task memory}: it emerges only
when accumulated artifacts degrade future performance through a
retrieval bottleneck.
Catastrophic forgetting in weight-update systems occurs when new
gradients overwrite previously learned
representations~\citep{kirkpatrick2017ewc}. Library drift is its
frozen-weight counterpart: persistent skill artifacts replace
neural weights as the degradation substrate, but the same
fundamental issue applies: accumulating new information can degrade
the system's prior performance unless lifecycle management actively
protects useful knowledge. Where EWC adds a regulariser over
parameter space, our fix adds retirement and a bounded cap over
skill space.

\paragraph{Diagnostic signals.}
LLM-as-judge~\citep{zheng2023llmjudge} reports $>$80\% agreement
with human preferences; we extend this into per-skill
\emph{attribution} verdicts (helped/hurt/neutral) from a closed
label set, requiring $\geq$3 failures sharing a canonical pattern
before any skill is born. This detects drift before it surfaces in
aggregate metrics, an early-warning system rather than a post-hoc
evaluator.

\section{Library Drift: Definition and Trigger}
\label{sec:failure}

\subsection{Operational definition}

Library drift occurs when a self-evolving skill library's
accumulated artifacts \emph{reduce} the agent's expected
performance below its no-skill baseline on a fixed task
distribution. Let $\mathcal{S}_t$ be the active skill set
at round $t$ and $p_0$ the no-skill pass rate. Drift manifests when:
\begin{equation}
  \mathbb{E}[\text{pass@1} \mid \mathcal{S}_t] < \mathbb{E}[p_0]
  \label{eq:drift}
\end{equation}
for some $t > 0$. Crucially, drift can occur even when every
individual skill appears reasonable; the failure is systemic,
arising from retrieval dilution and stale injection at the
\emph{library} level.

\subsection{Mechanism}

Library drift proceeds through three compounding stages:
\begin{enumerate}
  \item \textbf{Accumulation without quality signal.} Skills are
  born from failure patterns but never validated against outcomes.
  Marginal or harmful skills dilute the retrieval pool with each
  round.
  \item \textbf{Retrieval degradation.} The router selects skills
  by surface similarity. As the bank grows unbounded,
  near-duplicate or stale skills crowd out useful ones, degrading
  precision at constant recall.
  \item \textbf{Silent injection harm.} A stale skill injected
  into the solver prompt actively misleads, but produces no
  explicit error; the solver simply fails more often,
  indistinguishable from inherent task difficulty.
\end{enumerate}

In practice, these stages produce three distinguishable drift
sub-modes depending on the governance regime:
\emph{stagnation}: skills accumulate but fail to reach the solver
(no routing signal, no learning; A1 reproduces this floor);
\emph{bloat}: unbounded growth degrades retrieval until
injections become harmful (the failure mode of systems without a
cap~\citep{wang2023voyager, zhao2024expel}; prevented in our
Default by the active-cap); and \emph{erosion}: over-aggressive
governance destroys useful skills faster than they accumulate,
collapsing the bank (A4 retains only 2 active skills).
Which sub-mode dominates depends on governance strength: too
little permits bloat, too much causes erosion, and disconnecting
injection produces stagnation.

\subsection{Reproducible triggers}

We demonstrate library drift with two ablations of our full system
(Ratchet; Sec.~\ref{sec:fix}), both run on MBPP+ hard-100
\citep{liu2023evalplus} with Claude Opus 4.7 over 100 rounds:

\textbf{A1 (no skill injection):} The Router is forced to
\textsc{none}: the full pipeline (synthesizer, critic, curator)
still runs, but no skill is ever injected into the solver prompt.
The critic produces 0 calls because no skill-attributed failures
exist to judge.
Result: +0.002$\pm$0.005 gain (the no-skill floor). This
isolates the routing effect: skill creation alone, without
injection, produces no gain.

\textbf{A4 (harsh retirement):} Evidence floor $N_{\min}$ reduced
from 100 to 20; threshold $\tau$ tightened to 0.0. Result:
$-$0.019$\pm$0.010, \emph{below} the no-skill baseline. The
library actively harms performance. With only 20 trials, the
Hoeffding deviation is $\epsilon \approx 0.44$; skills with true
contribution $c \in [-0.44, 0]$ are prematurely retired on
unlucky draws, while genuinely harmful skills may survive early
rounds before enough evidence accumulates. The bank collapses to
2 active skills, and the router's 18.9\% engagement means it
rarely injects, yet when it does, the surviving skills hurt.
The effect is consistent across all three seeds ($-$0.005,
$-$0.027, $-$0.025; Table~\ref{tab:per-seed}), confirming that
harsh retirement reliably triggers erosion rather than reflecting
a single unlucky run.

Together, these triggers bracket the failure mode from both sides:
A1 shows that skills must be \emph{injected} to matter; A4 shows
that na\"ive lifecycle management can be \emph{worse} than none.
A4 is a cautionary negative result:
governance is not uniformly beneficial: acting on insufficient
evidence ($N_{\min}\!=\!20$ vs.\ the Default's 100) actively harms
performance rather than merely failing to help.

\section{Trace-Level Diagnostics}
\label{sec:diagnostics}

End-task metrics (pass@1, solve rate) are insufficient to diagnose
library drift: they conflate task difficulty, model stochasticity,
and library quality into a single number. We introduce three
complementary diagnostic signals that decompose the failure into
actionable components:

\subsection{Per-skill contribution score}

For each active skill $s$, we track:
\begin{equation}
  \hat{c}(s) = \frac{\text{successes}(s) - \text{failures}(s)}{\text{trials}(s)}
\end{equation}
computed from the evidence log (capsules where $s$ was injected).
A declining mean $\hat{c}$ across the bank signals drift: the
library is accumulating skills that do not help. In the Default
condition, mean $\hat{c}$ rises over rounds (skills that hurt are
retired); in A4, it oscillates because premature retirement removes
useful skills.

\subsection{Attribution verdicts}

For every failure capsule, a separate Critic LLM call produces a
structured verdict: \textsc{helped}, \textsc{hurt},
\textsc{neutral}, or \textsc{inapplicable}, plus a pattern label
and confidence. These verdicts serve dual purpose: (1)~synthesis
substrate (clusters of $\geq$3 failures sharing a canonical pattern
trigger new skills), and (2)~early-warning diagnostic. A rising proportion of
\textsc{hurt} verdicts is a leading indicator of drift: in A4,
the bank collapses to 2 active skills by round 30 (visible in
router engagement dropping to 18.9\%), while aggregate pass@1
declines only gradually because the router adaptively selects
\textsc{none} on most tasks.

\subsection{Router engagement metrics}

We track what fraction of tasks the router assigns a skill versus
\textsc{none}. Table~\ref{tab:main} shows that healthy
conditions maintain 70--80\% engagement, while A4 (drifting)
drops to 19\% as the bank empties. A2 (retrieval-only, no LLM
gate) shows 98\% engagement but lower quality; the LLM gate's
ability to decline injection is itself a drift-prevention mechanism.

\section{Verified Fix: The Ratchet Recipe}
\label{sec:fix}

Having defined the failure mode and the diagnostics that expose it,
we now describe the fix.
Ratchet is a single-agent loop where a frozen LLM writes,
retrieves, curates, and retires its own natural-language skills.
Three governance mechanisms jointly prevent library drift:

\subsection{Outcome-driven retirement}

A skill is retired once $n(s) \geq N_{\min}$ trials have
accumulated \emph{and} the empirical contribution
$\hat{c}(s) \leq -\tau$. The Default uses $N_{\min}=100$,
$\tau=0.10$, conservative enough that useful skills survive
stochastic noise (Hoeffding $\epsilon \approx 0.20$) but
aggressive enough that harmful skills are eventually removed.
This directly addresses the ``accumulation without quality signal''
stage of drift.

\subsection{Bounded active-cap}

A hard cap $C$ (Default: 50) forces the curator to evict the
lowest-contribution skill when synthesis would exceed the cap.
This directly addresses the ``retrieval degradation'' stage: by
bounding the bank, retrieval precision cannot decay with
unbounded growth. Combined with retirement, it provides a formal
non-divergence guarantee:

\textbf{Proposition 1.} Under bounded cap $C$ and retirement
threshold $\tau$, the expected eval pass@1 cannot drift below the
no-skill floor by more than $\tau + \epsilon + C\delta$, where
$\epsilon$ is the Hoeffding estimation tolerance and $\delta$ is
the per-skill failure probability.

Systems without bounded $C$ and $\tau$~\citep{wang2023voyager,
zhao2024expel, chen2024automanual} have no finite analogue of
this bound: library drift is unconstrained.

\subsection{Meta-skill authoring prior}

A meta-skill document constrains the Synthesizer to produce
stylistically consistent skills. This addresses the ``silent
injection harm'' stage at its source: by enforcing structural
homogeneity, fewer harmful or redundant skills are born in the
first place, reducing the load on downstream retirement. Ablation
A3 (no meta-skill) shows that removing it costs 43\% of the
Default's gain; it is the single most valuable component.

Surprisingly, explicit deduplication mechanisms (pattern
canonicalisation, A5; cover-guard, A6) are \emph{not} necessary
given the meta-skill: A5 and A6 slightly \emph{exceed} the
Default. The meta-skill subsumes explicit dedup at this scale, a
design insight relevant to any system with LLM-authored artifacts.

\section{Experiments}
\label{sec:experiments}

\subsection{Protocol}

We evaluate on \textbf{MBPP+ hard-100}: from the 378-task MBPP+
test split~\citep{liu2023evalplus}, we discard tasks that Claude
Opus 4.7 solves on all 5 baseline seeds ($\sim$273 tasks) and
randomly sample 100 from the remainder (60 train / 40 eval, fixed
seed). This isolates tasks where a skill library can plausibly
help. All LLM calls use Claude Opus
4.7; embeddings use Cohere embed-v4 (both via Amazon Bedrock). The Solver is a single direct
LLM call with no execution feedback, no self-refinement, and no
tool use, isolating the skill library's contribution. Each run is
100 rounds; we report mean $\pm$ std over 3 seeds.

\subsection{Main results}

\begin{table*}[t]
  \centering
  \small
  \caption{MBPP+ hard-100 (mean$\pm$std, 3 seeds).
  \emph{Baseline}: round-0 pass@1 (no skill active in any condition; cross-condition
  variation is sampling noise from $n\!=\!40$ eval tasks; gain is computed
  within each run, cancelling this).
  \emph{Gain}: mean(last 10) $-$ mean(first 10) of held-out pass@1.
  \emph{Router}: \% of eval tasks assigned a skill.
  \emph{Active/Retired}: bank state at round~100 (A1 has 42 active because the
  pipeline still synthesises skills, they are just never injected; A4's
  aggressive retirement empties the bank to 2, directly causing harm).
  \emph{Critic}: LLM verdict calls (0 = no quality signal).}
  \label{tab:main}
  \begin{tabular}{@{}lcccrrrr@{}}
    \toprule
    Condition & Baseline & Peak & Gain & Router & Active & Retired & Critic \\
    \midrule
    Default               & 0.258$\pm$0.047 & 0.658$\pm$0.042 & $+$0.328$\pm$0.018 & 73 & 50  &  89 & 4299 \\
    \midrule
    A1 no injection       & 0.283$\pm$0.031 & 0.375$\pm$0.000 & $+$0.002$\pm$0.005 &  0 & 42  &  15 &    0 \\
    A2 retrieval          & 0.242$\pm$0.012 & 0.492$\pm$0.042 & $+$0.077$\pm$0.065 & 98 & 42  &  69 & 5740 \\
    A3 no meta            & 0.200$\pm$0.035 & 0.592$\pm$0.047 & $+$0.187$\pm$0.036 & 80 & 50  &  84 & 4676 \\
    A4 harsh retire       & 0.300$\pm$0.035 & 0.433$\pm$0.042 & $-$0.019$\pm$0.010 & 19 &  2  &  51 & 1090 \\
    A5 no canon           & 0.275$\pm$0.020 & 0.708$\pm$0.012 & $+$0.374$\pm$0.023 & 80 & 50  &  76 & 4393 \\
    A6 no guard           & 0.217$\pm$0.024 & 0.700$\pm$0.035 & $+$0.363$\pm$0.033 & 70 & 50  &  94 & 3871 \\
    A7 cap=100            & 0.292$\pm$0.042 & 0.650$\pm$0.089 & $+$0.317$\pm$0.110 & 75 &100  &  55 & 4609 \\
    A8 meta refresh       & 0.250$\pm$0.035 & 0.725$\pm$0.020 & $+$0.372$\pm$0.017 & 74 & 50  & 131 & 4388 \\
    \bottomrule
  \end{tabular}
\end{table*}

\begin{figure*}[t]
  \centering
  \includegraphics[width=\textwidth]{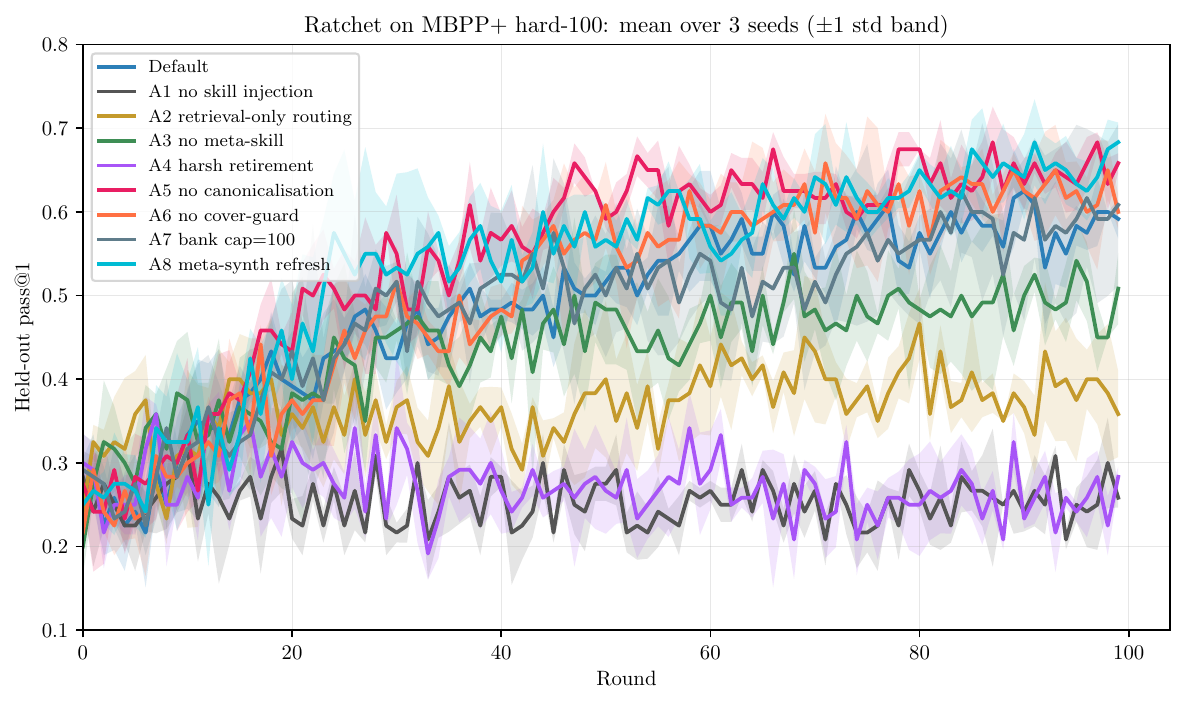}
  \caption{Held-out pass@1 by round (3-seed mean $\pm$1 std).
  A1 (flat floor) and A4 (below floor) exhibit library drift.
  A5/A6 (relaxed dedup) slightly exceed the Default; the meta-skill
  subsumes explicit filtering. A7 (doubled cap) shows comparable mean
  but higher variance. A8 (meta-synth refresh) matches A5/A6 gains
  but at 55\% more wall time (10.1\,h vs.\ 6.5\,h).}
  \label{fig:curves}
\end{figure*}

Table~\ref{tab:main} and Figure~\ref{fig:curves} summarize results.
The Default (full governance) lifts pass@1 from 0.258 baseline to
0.584 late-window mean (peak 0.658), more than doubling
performance on genuinely difficult tasks.
A1 establishes the no-skill floor (+0.002); A4 confirms that
drift can push the library \emph{below} it ($-$0.019).

\subsection{Decomposing the fix}

Eight ablations (A1--A8) each modify one knob from the Default to
test whether the corresponding mechanism is load-bearing
(3 seeds per condition; full settings in
Table~\ref{tab:hyperparams}). We group them into two questions.

\paragraph{A1--A3: which components are necessary?}
\textbf{A1} (no skill injection) forces the Router to always
select \textsc{none}: the synthesis pipeline still runs but no
skill reaches the solver. Result: $+$0.002 (the no-skill floor).
\textbf{A2} (retrieval-only routing) bypasses the LLM gate and
injects the top-ranked retrieval hit directly. Result: $+$0.077
(24\% of the Default gain), confirming the LLM gate contributes
beyond tf-idf$\cup$embed similarity.
\textbf{A3} (no meta-skill) removes the authoring prior from the
Synthesizer prompt. Result: $+$0.187 (57\% of the Default
gain), making the meta-skill the single most valuable component
($-$0.141 when removed).

\paragraph{A4: harsh retirement is harmful.}
A4 lowers the evidence floor $N_{\min}$ from 100 to 20 and
tightens $\tau$ to 0.0. Result: $-$0.019, \emph{below} the
no-skill floor. At $N_{\min}\!=\!20$, the Hoeffding deviation is
$\epsilon \approx 0.44$; skills with true contribution
$c \in [-0.44, 0]$ can be retired on unlucky draws, collapsing
the bank to 2 active skills. This validates Prop.~1: the
retirement threshold is insufficient without a sufficiently large
evidence floor. A4 is a cautionary result: governance is not
uniformly beneficial; acting on insufficient evidence actively
harms performance rather than merely failing to help.

\paragraph{A5--A6: explicit dedup is not necessary.}
A5 (no canonicalisation) raises $\tau_\text{canon}$ to 1.0,
disabling pattern dedup. A6 (no cover-guard) disables
duplicate-cluster skipping. Both slightly \emph{exceed} the
Default: A5 $+$0.374, A6 $+$0.363 (vs.\ Default $+$0.328;
within $\pm 2\sigma$ at $n\!=\!3$). The meta-skill's authoring
guidance enforces enough stylistic consistency that the explicit
filter's false positives discard more useful skills than
duplicates it prevents, a transferable design insight for any
system with LLM-authored artifacts.

\paragraph{A7--A8: relaxed cap and meta-skill refresh.}
A7 (bank cap=100) doubles the active-cap. Result: $+$0.317
comparable mean but substantially higher variance ($\pm$0.110
vs.\ $\pm$0.018); the cap primarily controls variance rather
than mean performance at this scale.
A8 (meta-synth refresh) regenerates the meta-skill every
10 rounds. Result: $+$0.372 and the highest peak (0.725), but at
55\% more wall time (10.1\,h vs.\ 6.5\,h) due to additional
synthesis rounds. Conclusion: more frequent refresh does not
meaningfully improve the learning curve but incurs substantial
compute overhead, not justified at this scale.

\subsection{Operational cost}

The Default uses $\sim$14.5k LLM calls per 100 rounds (10k solver
+ 4.3k critic + 152 synthesis), 43\% more than the no-skill
baseline (A1, 10k calls). Wall time is 6.5\,h vs.\ A1's
2.3\,h (2.8$\times$). The solver dominates cost in all
conditions (63--99\% of total calls).

\section{Discussion}
\label{sec:discussion}

\paragraph{Library drift as a general failure mode.}
Although we demonstrate drift in a skill library, neither the
definition (Eq.~\ref{eq:drift}) nor the diagnostics are specific
to skills. Rule-based systems~\citep{zhang2026ruleshaping},
workflow memories~\citep{wang2024awm}, and episodic stores all
persist LLM-authored artifacts across tasks and face the same
accumulation-without-retirement failure. We argue that library drift
deserves recognition as a first-class failure category alongside
tool-use errors and planning failures.

\paragraph{Diagnostics precede fixes.}
The trace-level diagnostics detected drift before end-task
metrics declined. In A4, the bank collapsed to 2 active skills by
round 30, but pass@1 showed only gradual decline because the
router adaptively chose \textsc{none} more often. Without
per-skill diagnostics, drift would appear as ``random task
difficulty variation,'' invisible and unfixable.
Critically, these diagnostics are not merely descriptive; they
enable counterfactual intervention. In the Default condition,
monitoring mean $\hat{c}$ and triggering retirement only after
$N_{\min}\!=\!100$ trials prevents the premature eviction that
destroys A4. An operator watching router engagement drop below
50\% could halt synthesis and investigate before drift reaches
end-task metrics.
The recipe (per-artifact contribution scores, attribution verdicts,
and engagement metrics) generalizes to any system where an LLM
agent accumulates persistent artifacts across episodes. We suggest
adopting these trace-level signals as a standard diagnostic
vocabulary for failure-mode analysis in agentic systems.

\paragraph{Expected vs.\ unexpected results.}
Of eight ablations, three produced surprises.
\emph{A4} fell below the no-skill baseline; we
expected harsh retirement to underperform, but not to actively
harm, confirming that the evidence floor is load-bearing even in
principle. \emph{A5/A6} slightly exceeded the Default, refuting
our design assumption that explicit dedup would help: the
meta-skill already enforces enough homogeneity. \emph{A8} yielded
only marginal gain despite 55\% more wall time, challenging the
intuition that a stale meta-skill limits synthesis quality. For
practitioners diagnosing drift in their own systems: governance
mechanisms can be counterproductive; the diagnostic recipe
(Sec.~\ref{sec:diagnostics}) is what makes failures visible and
safe intervention possible.

\paragraph{Scale-dependence of governance knobs.}
The Default is broadly safe: every condition except A4
substantially outperforms the no-skill baseline, and the
non-divergence guarantee (Prop.~1) holds regardless of task
distribution. However, exact knob settings are scale-dependent.
On our 100-task suite the meta-skill alone provides sufficient
stylistic consistency that canonicalisation (A5) and cover-guard
(A6) are unnecessary; on larger suites with hundreds of candidate
patterns, we expect explicit deduplication to become load-bearing.
Similarly, the bounded cap ($C\!=\!50$) is safe to double (A7
achieves comparable mean gain) but at the cost of substantially
higher variance: per-seed gains range from $+$0.172 to $+$0.440
(Table~\ref{tab:per-seed}), confirming that a relaxed cap exposes
the loop to luck-of-synthesis-order.
The principled approach: treat the Default as a safe starting
point, then relax governance knobs guided by the diagnostic
signals once the operator observes the meta-skill suffices.

\paragraph{Limitations.}
(i)~One benchmark (MBPP+ hard-100); generalization to broader
domains (multi-step agents, SWE-Bench) is left to future work.
(ii)~Single model (Claude Opus 4.7); cross-model stability not
established.
(iii)~The diagnostic thresholds (when to intervene) are empirically
chosen; principled calibration is future work.
(iv)~We hypothesize that drift signals (contribution scores,
\textsc{hurt} verdicts) become more sensitive in multi-step agents
where per-step verdicts can isolate which step a stale skill
harmed; empirical validation is future work.

\section{Conclusion}

The bottleneck in self-evolving skill libraries is not the
author; it is the librarian. Library drift, unbounded
accumulation without lifecycle management, is a silent, systemic
failure mode: the same frozen LLM that stagnates at $+$0.002
without governance delivers $+$0.328 once retirement, a bounded
cap, and an authoring prior are added. We provide (1)~reproducible
triggers that bracket the failure from both sides (A1: no
injection floor; A4: governance-induced harm), (2)~trace-level
diagnostics that detect drift before end-task metrics decline, and
(3)~eight ablations documenting which mechanisms are load-bearing
(retirement, meta-skill) and which are safely removed (explicit
dedup). Three surprises (A4 harmful, A5/A6 exceeding the Default,
A8 marginal despite cost) demonstrate that na\"ive governance can
be worse than none, making the diagnostic recipe essential for safe
intervention. The playbook (per-artifact contribution scores,
attribution verdicts, and engagement metrics) transfers to any
system that persists LLM-authored artifacts across episodes.

\section*{Impact Statement}

This paper presents work whose goal is to advance the field of
Machine Learning. There are many potential societal consequences
of our work, none which we feel must be specifically highlighted
here.

\newpage
\appendix

\section{System Architecture}
\label{app:architecture}

Ratchet operates a five-phase loop per round:
(1)~\textbf{Eval}: routes active skills on held-out tasks;
(2)~\textbf{Train}: same pipeline on train split, generating
failure substrate;
(3)~\textbf{Critic}: per-failure attribution verdict (LLM call);
(4)~\textbf{Synthesizer}: clusters failures by canonical pattern,
authors new skills from clusters with $\geq$3 members;
(5)~\textbf{Curator}: computes contribution scores, retires
under-performers, enforces cap.

The Router selects one skill or \textsc{none} per task via
two-stage retrieval (tf-idf + embedding, $K$=10 each) followed by
an LLM gate.

\section{Hyperparameters}
\label{app:hyperparams}

\begin{table}[h]
  \centering
  \small
  \caption{Default configuration hyperparameters.}
  \label{tab:hyperparams}
  \begin{tabular}{llr}
    \toprule
    Knob & Purpose & Value \\
    \midrule
    Active-cap $C$ & Max active skills & 50 \\
    $N_{\min}$ & Evidence floor & 100 \\
    $\tau$ & Retirement threshold & 0.10 \\
    Canon threshold & Pattern dedup & 0.85 \\
    Cover threshold & Cluster skip & 0.85 \\
    Bank-dedup threshold & Skill YAML dedup & 0.85 \\
    Skill length budget & YAML char cap & 1500 \\
    Synth lookback & Verdict window & 6 rounds \\
    Min cluster size & Synthesis trigger & 3 \\
    Max skills/round & Synthesis cap & 2 \\
    Router cutoff & Full-bank mode & 20 \\
    Retrieval $K$ & Shortlist size & 10 \\
    Rollback $\tau_\text{rb}$ & Regression depth & 0.10 \\
    Rollback persistence & Consecutive rounds & 5 \\
    Rounds & Per run & 100 \\
    Seeds & Per condition & 42, 7, 13 \\
    \bottomrule
  \end{tabular}
\end{table}

\section{Per-Seed Results}
\label{app:per-seed}

\begin{table}[h]
  \centering
  \small
  \caption{Per-seed rolling-mean gain and peak on MBPP+ hard-100.}
  \label{tab:per-seed}
  \begin{tabular}{llccc}
    \toprule
    Condition & Seed & Baseline & Peak & Gain \\
    \midrule
    Default          & 42 & 0.225 & 0.675 & $+$0.303 \\
                     & 7  & 0.225 & 0.600 & $+$0.340 \\
                     & 13 & 0.325 & 0.700 & $+$0.343 \\
    \midrule
    A1 no injection  & 42 & 0.325 & 0.375 & $+$0.003 \\
                     & 7  & 0.275 & 0.375 & $-$0.005 \\
                     & 13 & 0.250 & 0.375 & $+$0.008 \\
    \midrule
    A2 retrieval     & 42 & 0.250 & 0.475 & $+$0.148 \\
                     & 7  & 0.250 & 0.550 & $+$0.095 \\
                     & 13 & 0.225 & 0.450 & $-$0.010 \\
    \midrule
    A3 no meta       & 42 & 0.175 & 0.525 & $+$0.148 \\
                     & 7  & 0.250 & 0.625 & $+$0.177 \\
                     & 13 & 0.175 & 0.625 & $+$0.235 \\
    \midrule
    A4 harsh retire  & 42 & 0.325 & 0.450 & $-$0.005 \\
                     & 7  & 0.325 & 0.375 & $-$0.027 \\
                     & 13 & 0.250 & 0.475 & $-$0.025 \\
    \midrule
    A5 no canon      & 42 & 0.300 & 0.725 & $+$0.342 \\
                     & 7  & 0.275 & 0.700 & $+$0.385 \\
                     & 13 & 0.250 & 0.700 & $+$0.395 \\
    \midrule
    A6 no guard      & 42 & 0.200 & 0.725 & $+$0.365 \\
                     & 7  & 0.200 & 0.725 & $+$0.402 \\
                     & 13 & 0.250 & 0.650 & $+$0.322 \\
    \midrule
    A7 cap=100       & 42 & 0.350 & 0.525 & $+$0.172 \\
                     & 7  & 0.250 & 0.700 & $+$0.340 \\
                     & 13 & 0.275 & 0.725 & $+$0.440 \\
    \midrule
    A8 meta refresh  & 42 & 0.200 & 0.750 & $+$0.365 \\
                     & 7  & 0.275 & 0.700 & $+$0.355 \\
                     & 13 & 0.275 & 0.725 & $+$0.395 \\
    \bottomrule
  \end{tabular}
\end{table}

\section{Non-Divergence Proof Sketch}
\label{app:proof}

\textbf{Proposition 1.} Let $|\mathcal{S}_t| \leq C$, and suppose
the Curator retires skill $s$ when $n(s) \geq N_{\min}$ and
$\hat{c}(s) \leq -\tau$. Under Hoeffding bounds with tolerance
$\epsilon$ and per-skill failure probability $\delta$:

With probability $\geq 1-C\delta$ (union bound over $C$ skills),
every surviving skill has $c(s) \geq -\tau - \epsilon$. Tasks
routed to a surviving skill have expected pass probability
$\geq p_0(x) - \tau - \epsilon$; tasks routed to \textsc{none}
achieve $p_0(x)$. Taking expectations:
$\mathbb{E}[\text{pass@1}] \geq \mathbb{E}[p_0] - (\tau + \epsilon) - C\delta$.

With Default values ($\tau=0.10$, $N_{\min}=100$, $C=50$,
$\delta=10^{-3}$): $\epsilon \approx 0.20$, floor $=
\mathbb{E}[p_0] - 0.35$. The bound is loose at our scale (the
system gains +0.328 rather than losing 0.35) but rules out
unbounded degradation.

\end{document}